%% file: main.tex
\documentclass[10pt,twocolumn,letterpaper]{article}

\usepackage{cvpr}              

\usepackage{graphicx}
\usepackage{amsmath}
\usepackage{amssymb}
\usepackage{booktabs}
\usepackage{subcaption}
\usepackage{stmaryrd}
\usepackage{multirow}
\usepackage{enumitem}
\usepackage[dvipsnames,table,xcdraw]{xcolor}

\newcommand{\TN}[1]{{\color{black} \color{black}{}#1}}
\newcommand{\ST}[1]{{\color{black} \color{black}{}#1}}

\newcommand{\KGnew}[1]{{\color{black} \color{black}{}#1}}
\newcommand{\STnote}[1]{{\color{blue}{} \color{blue}{}}}
\newcommand{\KGnote}[1]{{\color{red}{} \color{red}{}}}
\newcommand{\cc}[1]{{\color{magenta} \color{magenta}{}}}
\newcommand{\TNnote}[1]{}

\newcommand{\MODELNAME}{EgoDistill}
\newcommand{\EPK}{EPIC-Kitchens}
\newcommand{\aprox}{\raisebox{0.5ex}{\texttildelow}}

\usepackage[pagebackref,breaklinks,colorlinks]{hyperref}

\usepackage[capitalize]{cleveref}
\crefname{section}{Sec.}{Secs.}
\Crefname{section}{Section}{Sections}
\Crefname{table}{Table}{Tables}
\crefname{table}{Tab.}{Tabs.}

\def\cvprPaperID{6959} 
\def\confName{CVPR}
\def\confYear{2023}

\begin{document}

\title{\MODELNAME: Egocentric Head Motion Distillation \\for Efficient Video Understanding}

\author{
	Shuhan Tan$^{1}$\quad Tushar Nagarajan$^{1}$ \quad Kristen Grauman$^{1,2}$\\
	$^{1}$The University of Texas at Austin \quad $^{2}$FAIR, Meta AI \\
    {\tt\small \hspace{0mm}\{shuhan,tushar.nagarajan,grauman\}@cs.utexas.edu}
}

\maketitle

\begin{abstract}
\input{latex/sections/abstract}
\end{abstract}

\section{Introduction}
\label{sec:intro}
\input{latex/sections/introduction}

\section{Related Work}
\label{sec:related}
\input{latex/sections/related_works}

\section{Approach} 
\label{sec:method}
\input{latex/sections/method}

\section{Experiments}
\label{sec:exp}
\input{latex/sections/experiment}

\section{Conclusion}
\label{sec:con}
\input{latex/sections/conclusion}

{\small
\bibliographystyle{ieee_fullname}
\bibliography{cites}
}

\newpage

\input{latex/supp.tex}

\end{document}

%% file: latex/sections/abstract.tex
Recent advances in egocentric video understanding models are promising, but
their heavy computational expense
is a barrier for many real-world applications.
To address this challenge,
we propose \MODELNAME, a distillation-based approach that learns to reconstruct heavy egocentric video clip features by combining the semantics from a sparse set of video frames with the head motion from lightweight IMU readings.
We further devise a novel self-supervised training strategy for IMU feature learning.
Our method leads to significant improvements in efficiency, requiring $200 \times$ fewer GFLOPs than equivalent video models. We demonstrate its effectiveness on the Ego4D and EPIC-Kitchens datasets, where our method outperforms state-of-the-art efficient video understanding methods. 

%% file: latex/sections/introduction.tex
Recent advances in augmented and virtual reality (AR/VR) technology have the potential to change the way people interact with the digital world, much like the smartphone did in the previous decade.
A fundamental requirement for AR/VR systems is the ability to recognize user behavior from egocentric video captured from a head-mounted camera.
Towards this goal, several egocentric video datasets have been proposed in recent years, spurring increasing attention of the research community~\cite{Ego4D2022CVPR, sigurdsson2018charades,Damen2022RESCALING}.

Recent advances in egocentric action recognition, anticipation, and retrieval focus on building powerful \emph{clip-based} video models that operate on video clips of a few seconds \KGnew{at a time}~\cite{9008780, patrick2021keeping, girdhar2022omnivore, fan2021multiscale, Datta_2022_CVPR, Plizzari_2022_CVPR, Li_2022_CVPR, Liu_2022_CVPR}.
Despite
encouraging performance, these models typically process densely-sampled frames with temporally-aware operations, making them computationally heavy.
This makes them impractical for AR/VR devices with constrained resources, or for real-time video applications that require low latency.
How to efficiently perform egocentric 
\ST{video understanding}
is therefore an important\TN{, yet} unsolved problem.
\input{latex/figures/teaser}

To address this issue,
we take inspiration from how animals perceive the world with ego-motion.
Neuroscience research has found that during active movement, the animal visual cortex receives and utilizes head motion signals from the motor cortex for visual processing~\cite{GUITCHOUNTS2020512, PARKER2020581, Parker2022.02.01.478733}.  This indicates that head motion signals support an embodied agent's \ST{efficient} understanding of the egocentric visual stream.
Inspired by this phenomenon, we explore the relationship between human head motion and egocentric video for efficient video understanding.
In practice, we consider head motion signals captured by the inertial measurement unit (IMU) of a head-mounted camera. IMU measures motion from an accelerometer and gyroscope and is widely available on 
popular wearable devices. 
Prior work leverages IMU as an extra modality for human action recognition~\cite{s20102905, 8994060, 8730690},
(e.g., jumping, walking, standing) or as geometric cues for visual-inertial odometry~\cite{Cao_2022_CVPR, 7557075, Yang2022EfficientDV}.

In contrast, we propose to achieve efficient video understanding by drawing on IMU as a \emph{substitute} for dense video frame observations.  
The intuition is as follows. A video clip contains two things: semantic content (appearance of objects, places, people) and dynamics (how the scene and the camera move). 
While densely sampled frames are sure to capture both of the above---as done
by current clip models~\cite{patrick2021keeping, fan2021multiscale, Feichtenhofer2020X3DEA}---we hypothesize they are sometimes overkill.
For a short video clip, much of the semantic content is intelligible from even a single frame; meanwhile, the head motion provides a good portion of the dynamics,  implicitly revealing how the visual appearance changes across neighboring frames.

Building on this insight, we introduce \MODELNAME, an approach that learns
to reconstruct dense egocentric video clip features using temporally sparse visual observations (as few as one RGB frame) together with the head motion from IMU.
Specifically, \MODELNAME~employs a new form of knowledge distillation from video models.
During training, we train a lightweight model that takes sparsely sampled image(s) and IMU to approximate the video representation extracted by a powerful but expensive video model.
We further improve the model with a novel self-supervised training stage for IMU feature learning.
During inference, we directly utilize the lightweight model for egocentric video recognition, leading to much higher efficiency.
Our model is flexible to the target heavy video feature, as we demonstrate with multiple current leading 
egocentric video
models~\cite{patrick2021keeping, fan2021multiscale, 9008780, Feichtenhofer2020X3DEA}. See Figure~\ref{fig:teaser}.

Importantly, \MODELNAME~offers a major efficiency gain: processing low-dimensional IMU and a few frames is much more efficient compared to processing a dense stack of frames. In practice, \MODELNAME~uses $200 \times$ fewer GFLOPs than the original video model.

We experiment on the largest available 
egocentric action recognition datasets: Ego4D~\cite{Ego4D2022CVPR} and EPIC-Kitchens-100~\cite{Damen2022RESCALING}. We show that IMU coupled with an image offers better cross-modality knowledge distillation performance than images alone or images with audio.
For a typical 50-minute egocentric video, \MODELNAME~reduces inference time of the source video model from 25 minutes to \emph{36 seconds}.
Moreover, with only 1-4 frames, our lightweight
distillation model achieves a better accuracy-efficiency trade-off than state-of-the-art  
models for adaptively sampling video content~\cite{meng2021adafuse, wang2021efficient}.
Notably, we surpass the accuracy of these fast approaches by a large margin while requiring 4$\times$ to 8$\times$ less computation.

%% file: latex/figures/teaser.tex
\begin{figure}[t]
  \centering
    \includegraphics[width=\linewidth]{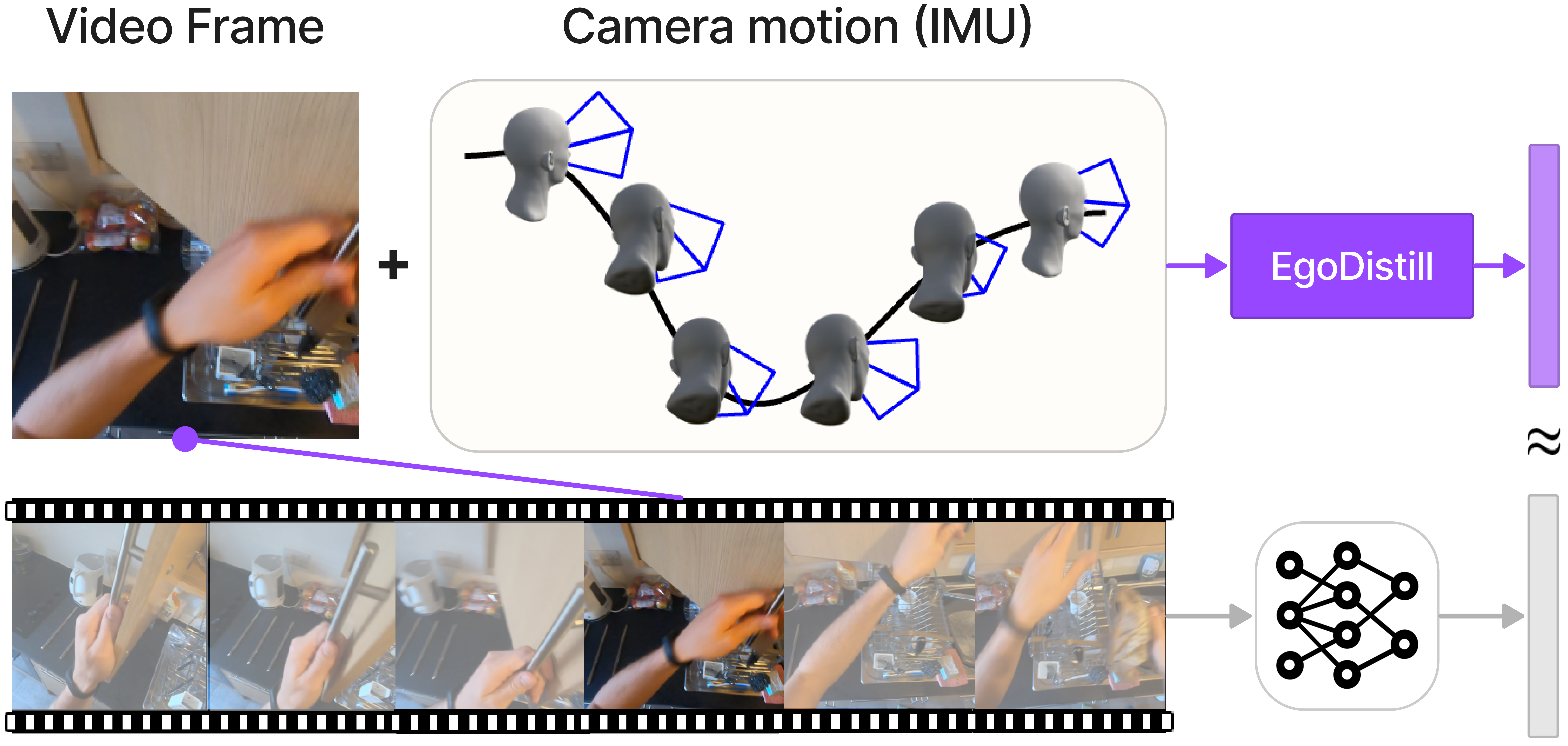}
    \caption{\textbf{Illustration of \MODELNAME.} Given a single video frame and camera motion from IMU, \MODELNAME~learns to reconstruct the more expensive dense video clip feature. With its lightweight input, \MODELNAME~significantly improves efficiency.
    }
   \label{fig:teaser}
\vspace{-0.3em}
\end{figure}

%% file: latex/sections/related_works.tex
\noindent \textbf{IMU for activity recognition.}
Recent work
explores using the IMU sensor on mobile devices for human activity recognition of actions like walking, jumping, or sitting~\cite{9680027, 9744439, Tao2021AttentionBasedSF, app10124213, 10.1145/3494995}. Normally, these models take input from IMU sensors mounted on human body joints~\cite{Tao2021AttentionBasedSF, 7350781, 10.1371/journal.pone.0075196}, waist-mounted~\cite{10.1145/1964897.1964918} or in-pocket smartphones~\cite{IGNATOV2018915}.
See~\cite{Wang2019DeepLF} for a survey.
Abundant work in video recognition explores
ways to learn from RGB coupled with other modalities---audio~\cite{Arandjelovi2017LookLA,gao2020listentolook, 10.5555/3327757.3327874}, optical flow~\cite{NIPS2014_00ec53c4, Sun_2018_CVPR, Feichtenhofer2016ConvolutionalTN} or both~\cite{panda2021adamml, kazakos2019TBN, huang2021what}---but
comparatively fewer use IMU~\cite{9187232}, and unlike our work, they focus on third-person video~\cite{s20102905, 8994060, 8730690} and do not target at model efficiency.
Our idea is for IMU to help reconstruct
more expensive video features, rather
than simply fuse IMU with RGB for multi-modal recognition.

\noindent \textbf{IMU for odometry.}
Inertial odometry aims to estimate the position and orientation of the camera-wearer with readings from the IMU sensor. Traditionally, methods rely on IMU double integration~\cite{Bortz1971ANM}
or enhancements thereof~\cite{inproceedings, Brajdic2013WalkDA, 5286542}.
Recent data-driven methods automatically learn to perform inertial odometry with supervised~\cite{9196860, Yan2018RIDIRI} or self-supervised learning~\cite{Cao_2022_CVPR}, or combine IMU and visual input for more robust estimates with visual-inertial odometry~\cite{7557075, Yang2022EfficientDV}. 
While IMU can convey geometric ego-motion
to our learned model, our goal is to \TN{produce} efficient egocentric video features rather than to output odometry.

\noindent \textbf{Visual feature learning with IMU.} 
IMU is also used to learn better vision features~\cite{jayaraman-iccv2015,ehsani2018dog, ehsani2020learning, Tsutsui2021HowYM}, 
e.g., \TN{to encourage} image features that are equivariant with ego-motion~\cite{jayaraman-iccv2015},  to predict an IMU-captured body part (leg, hand)~\cite{ehsani2018dog, ehsani2020learning}, or \TN{to} predict video-IMU correspondence~\cite{Tsutsui2021HowYM}, \TN{for} applications like action recognition~\cite{ehsani2020learning, Tsutsui2021HowYM} and scene understanding~\cite{ehsani2018dog, jayaraman-iccv2015}.
While these results reinforce that IMU can inject embodied motion into visual features, our idea to use 
head motion to infer pretrained video features for speedy video understanding is distinct.

\noindent \textbf{Efficient video recognition.} 
Being crucial for mobile applications, efficient
video recognition has received increasing attention in recent years.
\ST{Several studies focus on designing lightweight architectures~\cite{9196860, Feichtenhofer2020X3DEA, kondratyuk2021movinets, ECO_eccv18, Tran2018ACL} by reducing 3D CNN operations across densely sampled frames. 
Our idea is orthogonal to them as we focus on inputs with sparsely-sampled frames. 
\KGnew{As we} show in experiments, our method is compatible with different video architectures.}

\ST{Another line of research achieves efficiency by adaptively selecting video content to process.}
Some reduce \emph{temporal redundancy} by
adaptively selecting which video clip~\cite{Korbar2019SCSamplerSS}, frames~\cite{meng2020ar, ghodrati2021}, and/or feature channel~\cite{meng2021adafuse} to process and which to skip,
while others reduce \emph{spatial redundancy}, efficient recognition by dynamically selecting 
selecting for each frame a smaller but important region to process~\cite{wang2021adafocus, wang2022adafocusv2}.
Other work \KGnew{dynamically selects} tokens in video transformers among both the spatial and temporal dimensions~\cite{wang2021efficient}.
Our idea is complementary: rather than dynamically subsample the available video content, we show how to \emph{infer ``full" video features for every clip} using static image(s) and motion data.  Our results \KGnew{outperform} state-of-the-art sampling models (cf.~Sec.~\ref{sec:exp}).  In addition, we focus on egocentric video, where  head motion is particularly meaningful for inferring unobserved visual content. To our knowledge, ours is the first technique
specifically aimed at accelerating egocentric video processing.

\noindent \textbf{Multimodal distillation.}
Knowledge distillation aims to transfer knowledge learned by an expensive model to a lightweight model~\cite{Hinton2015DistillingTK}.
Recent work explores multimodal distillation, e.g.,
transferring 
from a RGB model to a flow or depth model~\cite{Garcia_2018_ECCV, 7780678}, from a 3D model to a 2D model~\cite{Liu20213Dto2DDF}, or from a visual model to audio model~\cite{10.5555/3157096.3157196, 9009049}.  
The ListenToLook model~\cite{gao2020listentolook} incorporates both clip subsampling and video-to-audio distillation for fast activity recognition in third-person video.
In contrast, we explore the relationship between the camera-wearer's head motion and RGB signals for egocentric video. Our experiments show \MODELNAME's advantage over ListenToLook in terms of the speed-accuracy tradeoff on egocentric video datasets.

%% file: latex/sections/method.tex
We introduce \MODELNAME, which uses sparsely-sampled frames and head motion from IMU to approximate the features of heavy video models for efficient egocentric video understanding.
We first introduce the egocentric action recognition task (Sec.~\ref{sec:action_rec}). Then, we introduce our pipeline (Sec.~\ref{sec:imu_action_rec}), our distillation model and training objective (Sec.~\ref{sec:distillation_model}), and our self-supervised IMU feature learning (Sec.~\ref{sec:pretrain}).
Figure~\ref{fig:method} overviews our approach.

\input{latex/figures/method}

\subsection{Egocentric action recognition} \label{sec:action_rec}
Given a fixed-length video clip $\mathcal{V} \in \mathbb{R}^{T \times H \times W \times 3}$ consisting of $T$ RGB frames of size $H \times W$ and a set of $C$ action classes, the task of action recognition is to output a score for each action class, representing its likelihood. Typically, this is done with a powerful but expensive video model $\Omega$, that directly operates on all the available frames to output
the $C$ class logits
$\Omega(\mathcal{V}) \in \mathbb{R} ^{C} $. $\Omega$ is trained with standard classification loss:
\begin{equation}
    \mathcal{L}_{\text{ACT}} = \sum_{\mathcal{V}_i} \mathcal{L}_{\text{CE}}( c_{i},  \sigma(\Omega(\mathcal{V}_i))),
    \label{eq:video}
\end{equation}
where $\mathcal{V}_{i}$ is the $i$-th video clip in the dataset, $c_{i}$ is the corresponding ground-truth action label, $\sigma$ is the softmax function, and $\mathcal{L}_{\text{CE}}$ is cross-entropy loss. 
Popular video recognition models use clips that are typically \aprox 2 seconds long~\cite{patrick2021keeping, fan2021multiscale, 9008780}. 
For longer videos, scores are averaged across all clips it contains to infer the video action label.

\subsection{Efficient video inference with head motion} \label{sec:imu_action_rec}
Processing the video clip $\mathcal{V}$ for action recognition is computationally intensive; however, the computation cost can be modulated depending on how frames from the clip are used. On the one hand, \emph{clip-based} models~\cite{9008780, patrick2021keeping,fan2021multiscale, Feichtenhofer2020X3DEA} process most (or all) frames in a video clip $\mathcal{V}$ to achieve strong recognition performance, but come at a high computational cost. On the other hand, \emph{frame-level} models~\cite{meng2020ar, ghodrati2021, panda2021adamml, wang2022adafocusv2} only process one (or a small number) of frames from $\mathcal{V}$ and are more efficient, but suffer a drop in performance as a result. 
Our goal is to train a frame-based model that can approximate heavy clip-based model performance while maintaining high efficiency.

For this, we turn to head motion captured by IMU.
Along with RGB frames, each video clip is paired with IMU measurements $\mathcal{M}$ that record the camera (head) motion during the video.
Specifically, the IMU readings are composed of 6-dimensional accelerometer and gyroscope measurements in the $xyz$ axes, which encode strong temporal motion information about camera pose changes (both translation and rotation) across frames. 

For short video clips, a set of sparsely sampled frames $\mathcal{I}$ often already captures most \emph{appearance} information.
Complementary to this, the IMU readings capture \emph{camera motion} information (\KGnew{see below for discussion on scene motion}).
Moreover, IMU is very efficient to process due to its low dimensionality.
By processing inputs from these two sources with a lightweight frame-based model, we can infer the semantic and dynamic features of a heavier clip-based video model. 

Given $\mathcal{I}$ and $\mathcal{M}$, we train an efficient lightweight model $\Phi$ to approximate the output of video model $\Omega$.  Specifically, we train our 
\MODELNAME~model $\Phi$ that achieves 
\begin{equation}
    \Phi(\mathcal{I}, \mathcal{M}) \approx \Omega(\mathcal{V}).
\label{video_appr}
\end{equation}
Such a lightweight model will be able to approximate the result of the heavy video model, while being much more efficient. Our approach is agnostic to the specific video model $\Omega$; in experiments, we demonstrate its versatility for MotionFormer~\cite{patrick2021keeping}, MViT~\cite{fan2021multiscale}, SlowFast~\cite{9008780} and X3D~\cite{Feichtenhofer2020X3DEA}. 

In practice, we uniformly sample $N$ frames\footnote{Other frame sampling heuristics (e.g., selecting from the start or center of the video) performed equivalently or worse than uniform sampling.}
from $\mathcal{V}$ to obtain $\mathcal{I}$. We can achieve a trade-off between efficiency and performance by changing the number of frames $N$. In our experiments we use very low values of $N$ (1 to 4 frames). 
In the next section, we discuss how we train $\Phi$. 

\subsection{Video feature distillation with IMU} \label{sec:distillation_model}
We achieve the objective in Equation~\ref{video_appr} via knowledge distillation~\cite{Hinton2015DistillingTK}, where 
we 
transfer knowledge learned by the expensive teacher model $\Omega$ to a lightweight student model $\Phi$.
Next we present the design of $\Phi$ and the training objectives, followed by our self-supervised IMU feature pretraining stage in Sec.~\ref{sec:pretrain}.

We design $\Phi$ to be a two-stream model. 
For a video clip and associated IMU signal $(\mathcal{I}, \mathcal{M})$, we extract image features $\mathbf{z}_{\mathcal{I}} = f_{\mathcal{I}}(\mathcal{I})$ and IMU features $\mathbf{z}_{\mathcal{M}} = f_{\mathcal{M}}(\mathcal{M})$ using lightweight feature encoders $f_{\mathcal{I}}$, $f_{\mathcal{M}}$ respectively.
Then, we fuse $\mathbf{z}_{\mathcal{I}}$ and $\mathbf{z}_{\mathcal{M}}$ with a fusion network $\Pi$ to obtain the fused VisIMU feature $\mathbf{z}_{\phi}  = \Pi(\mathbf{z}_{\mathcal{I}}, \mathbf{z}_{\mathcal{M}})$. Finally, a fully-connected layer uses the fused feature to predict class logits $\Phi(\mathcal{I}, \mathcal{M}) \in \mathbb{R} ^C $. 

The fused feature $\mathbf{z}_{\phi}$ contains semantic information from the image frame 
coupled with complementary motion information from IMU, allowing us to accurately reconstruct the video clip feature. 
\KGnew{See Figure~\ref{fig:method}.}

We train $\Phi$ with a combination of three losses, as follows. 
First, we train $\Phi$ to approximate the original video feature $\mathbf{z}_{\mathcal{V}}$ from the video model $\Omega$: 
\begin{equation}
    \mathcal{L}_1 = \sum_{(\mathbf{z}_{\mathcal{V}_i}, \mathbf{z}_{\phi_i})} \left\| \mathbf{z}_{\mathcal{V}_i} - \mathbf{z}_{\phi_i} \right\|_1.
\end{equation}
This cross-modal loss encourages the 
fused feature $\mathbf{z}_{\phi}$ to match the video feature,
i.e., the combined features from the different modalities should match in the feature space. 

Training with $\mathcal{L}_1$ alone does not fully capture the classification output of $\Omega$.
Therefore, we also train $\Phi$ with a knowledge distillation loss:
\begin{equation} 
    \mathcal{L}_{\text{KD}} = \sum_{(\mathcal{V}_i, \mathcal{I}_i, \mathcal{M}_i)} 
    \mathcal{D}_{\text{KL}} (\sigma(\Omega(\mathcal{V}_i)/\tau), \sigma(\Phi (\mathcal{I}_i, \mathcal{M}_i)/\tau)),
\end{equation}
where $(\mathcal{V}_i, \mathcal{I}_i, \mathcal{M}_i)$ represents the $i$-th clip in the dataset, $\mathcal{D}_{\text{KL}}$ measures KL-divergence between the class logits from the teacher model $\Omega$ and student model $\Phi$, and $\tau$ is a temperature parameter.
Intuitively, $\mathcal{L}_{\text{KD}}$ casts
the output of the video teacher model as a soft target for training the student model. In this way, the student model learns to better generalize by mimicking the output distribution of the heavy video model.

Finally, to further encourage the features to preserve elements useful for activity understanding, we also compute an action classification loss: 
\begin{equation}
    \mathcal{L}_{\text{GT}} = \sum_{(\mathcal{I}_i, \mathcal{M}_i)} 
    \mathcal{L}_{\text{CE}}( c_{i},  \sigma(\Phi (\mathcal{I}_i, \mathcal{M}_i))),
    \label{eq:gt}
\end{equation}
where $c_i$ is the ground-truth action label, following Equation~\ref{eq:video}. The final training loss is a combination of these three loss functions:
\begin{equation}
    \mathcal{L} = \alpha \mathcal{L}_{\text{KD}} + (1-\alpha) \mathcal{L}_{\text{GT}} + \beta \mathcal{L}_1,
    \label{eq:full}
\end{equation}
where $\alpha$ controls the balance between knowledge distillation and
activity training~\cite{Hinton2015DistillingTK}, and $\beta$ controls the weight for feature space matching. 

Critically, processing a few image frame(s) and the low-dimensional IMU readings is substantially faster than processing the entire video.
Once trained, our
model can approximate the behavior of the source video model for recognition tasks, with the key benefit of efficient egocentric video recognition. 

What kind of motion does our model preserve? Video motion decomposes into \emph{scene} motion  (e.g., how the objects and the camera wearer's hands are moving on their own), and \emph{camera} motion (i.e., how the camera wearer is moving their head). By itself, IMU would directly account only for camera motion, not scene motion. However, by learning to map from the RGB frame \emph{and} IMU to the \emph{full} video feature, we are able to encode predictable scene motions tied to scene content, e.g., how does hand and object movement in subsequent frames relate to the camera wearer's head motion (see Figure~\ref{fig:quali_scene_motion}). Moreover, our model is applied to relatively short clips (1-2 seconds) in sequence, which means the \KGnew{appearance} content is regularly refreshed as we \KGnew{slide down to} process the longer video. 

\subsection{\ST{Self-supervised IMU feature learning}}\label{sec:pretrain}
The success of \MODELNAME~depends on how well the IMU feature encoder $f_{\mathcal{M}}$ extracts useful camera motion information and associates it with the visual appearance change in the video clip.
\ST{In this way \MODELNAME~can learn to anticipate unseen visual changes in the video with $\mathcal{I}$ and $\mathcal{M}$.}
We design a self-supervised pretraining task to initialize the weights of $f_{\mathcal{M}}$ to achieve this.

Specifically, for each clip $\mathcal{V}$, we obtain its first and last frames $(\mathcal{I}^0, \mathcal{I}^T)$ 
as well as the IMU $\mathcal{M}$. We first extract visual features $\mathbf{z}_{\mathcal{I}}^0, \mathbf{z}_{\mathcal{I}}^T$ and IMU feature $\mathbf{z}_{\mathcal{M}}$ with feature extractors $f_{\mathcal{I}}$ and $f_{\mathcal{M}}$ mentioned above. Then, we train 
a feature predictor $h$ to predict the IMU feature $\hat{\mathbf{z}}_{\mathcal{M}} = h(\mathbf{z}_{\mathcal{I}}^0, \mathbf{z}_{\mathcal{I}}^T)$.
By connecting $\hat{\mathbf{z}}_{\mathcal{M}}$---which is a function of image features only---with $\mathbf{z}_{\mathcal{M}}$, we encourage $f_{\mathcal{M}}$ to extract useful camera motion features specifically associated with the visual appearance changes. 
Note that those appearance changes may include scene motion. 
\ST{
Therefore, we \KGnew{include an $\mathcal{L}_1$ loss to train $f_{\mathcal{M}}$}, which encourages $f_{\mathcal{M}}$ to extract motion features accounting for scene motion in the full video.
}

In sum, we train $f_{\mathcal{M}}$, $h$, and the fusion network $\Pi$ using $\mathcal{L}_1$ and NCE loss~\cite{pmlr-v9-gutmann10a}: $   \mathcal{L}_{\text{pretrain}} = \mathcal{L}_{\text{NCE}} + \mathcal{L}_1$, where
\begin{equation}
    \mathcal{L}_{\text{NCE}} = \sum_{i} - \log{ \frac{\text{sim}(\hat{\mathbf{z}}_{\mathcal{M}_i}, \mathbf{z}_{\mathcal{M}_i}) }{\sum_j \text{sim}(\hat{\mathbf{z}}_{\mathcal{M}_i}, \mathbf{z}_{\mathcal{M}_j})}}.
\end{equation}
We sample negative examples $\mathbf{z}_{\mathcal{M}_j}$ from other instances in the same mini-batch for $j \neq i$, and $\text{sim}(q,k) = \exp(\frac{q \cdot k}{|q||k|} \frac{1}{\tau'})$ 
with temperature $\tau' = 0.1$\footnote{
\TN{We keep the ImageNet-pretrained $f_{\mathcal{I}}$ model frozen, as finetuning it leads to mode collapse.}
}.

To summarize, prior to the main training stage of Equation~\ref{eq:full}, we pretrain the IMU feature extractor $f_{\mathcal{M}}$ and fusion network $\Pi$.
As we will show below, both pretraining losses result in IMU features that are consistent with visual changes and lead to better finetuning performance. 

%% file: latex/figures/method.tex
\begin{figure*}[ht]
  \centering
  \includegraphics[width=1.0\linewidth]{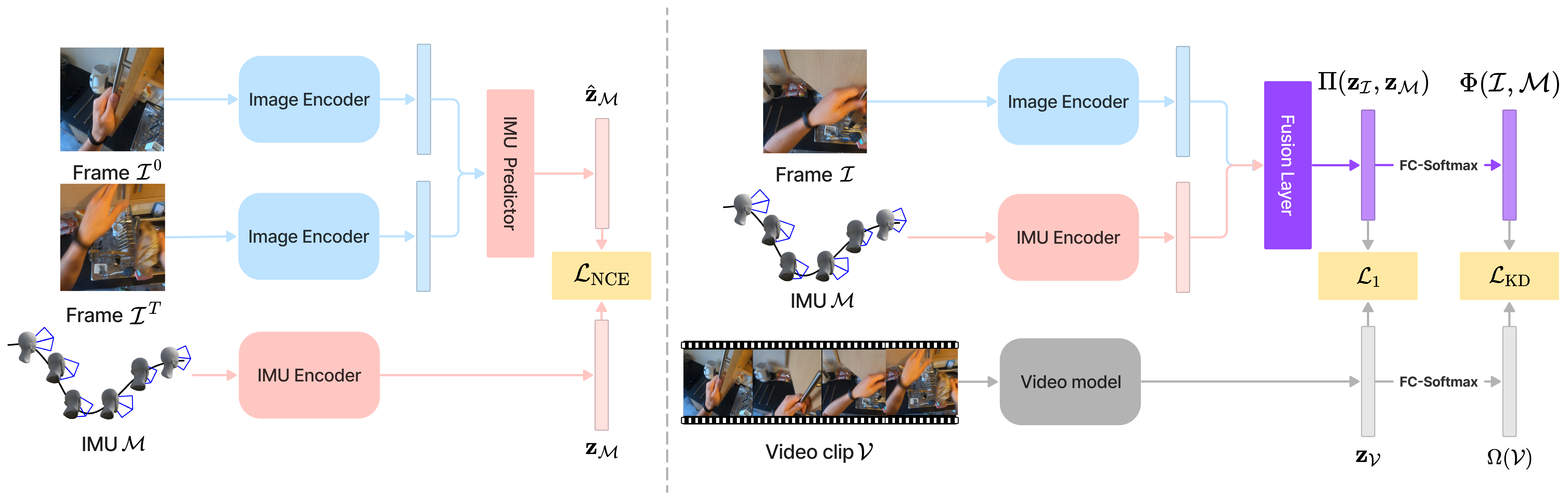}
  \vspace{-1.8em}
   \caption{\textbf{\MODELNAME~architecture.}
   \textbf{Left: Self-supervised IMU feature learning.} Given start and end frames of a clip, we train the IMU encoder to anticipate visual changes. \textbf{Right: Video feature distillation with IMU.} Given image frame(s) and IMU, along with our pre-trained IMU encoder, our method trains a lightweight model with knowledge distillation to reconstruct the features from a heavier video model.
  \ST{When \TN{the} input includes more than one image frame, the image encoder aggregates frame features temporally with a \KGnew{GRU}.}}
   \label{fig:method}
  \vspace{-1.0em}
\end{figure*}

%% file: latex/sections/experiment.tex
\input{latex/figures/main_curve}

We evaluate our approach for resource-efficient action recognition.

\subsection{Experimental setup}
\textbf{Datasets.}
We experiment on two large-scale egocentric action recognition datasets. 
\textbf{Ego4D}~\cite{Ego4D2022CVPR} contains 3,670 hours of egocentric videos of people performing diverse tasks (from cooking to farming) across the globe. As action recognition is not part of the original 
Ego4D benchmark, we construct this task with annotations from the Hands+Objects temporal localization benchmark~\cite{Ego4D2022CVPR} (see Supp.~for details). We include clips with paired IMU and audio\footnote{We require audio to compare with the audio-based baseline~\cite{gao2020listentolook}.}, and consider classes with at least 2 labeled instances.  
This results in a 94-class action recognition dataset with 8.5k training videos and 3.6k evaluation videos. 
\textbf{\EPK}~\cite{Damen2022RESCALING} contains 100 hours of egocentric videos capturing daily activities in kitchen environments. We use annotations from the action recognition benchmark. Similar to Ego4D, we select videos that have paired IMU and audio data, and split the resulting data by camera-wearer.
This results in a 62-class action dataset with 29k training videos and 6.2k evaluation videos. 
For both datasets, we use ``verb'' labels as the target for action recognition as they are well aligned to activity motions.

\textbf{Evaluation metrics.}
To measure action recognition performance, we report the per-video top-1 accuracy 
on the validation 
set.
We densely sample clips from each video and average their predictions to compute accuracy. 
To benchmark efficiency, we measure computational cost with FLOPs (floating-point operations) during inference.

\textbf{Implementation details.}
In our main experiments, we use MotionFormer~\cite{patrick2021keeping} as the video teacher model $\Omega$ due to its strong performance for egocentric video. For \EPK, we use the authors' 
provided checkpoint. For Ego4D, we finetune the above model for 50 epochs with $1e^{-4}$ learning rate and 64 batch size on the training set. 
We use 16-frame input with sample rate 4. 
For the student model $\Phi$, we use a ResNet-18 as the image backbone $f_{\mathcal{I}}$ and a 1D Dilated CNN~\cite{brossard2020denoising} for the IMU backbone $f_{\mathcal{M}}$. 
The feature fusion module $\Pi$ uses a concatenation operation following a two-layer fully-connected layer with hidden dimension 1024.
For each video clip, the input image(s) is resized to $224 \times 224$, and the IMU is a $422\times6$ matrix (around 2 seconds with 198Hz frequency), representing the accelerometer and gyroscope readings \TN{along the} $xyz$ axes.
For the image input, we uniformly sample $N$ frames from the video clip.
If $N>1$, we use $f_{\mathcal{I}}$ to sequentially generate features for each frame and aggregate them with a GRU module~\cite{Cho2014LearningPR}.
For both datasets, we first pretrain the model with the self-supervised objective \TN{(Section~\ref{sec:pretrain})}
for 50 epochs with AdamW~\cite{loshchilov2018fixing} using batch size 64 and learning rate $1e^{-4}$. Then, we finetune all the models with the same setting (Equation~\ref{eq:full}). 
We set $\alpha = 0.95$ and $\beta = 1.0$ based on validation data.  For Ego4D, we set $\tau=10.0$ and train the model for 150 epochs. For \EPK, we set $\tau=1.0$ and train for 50 epochs.

\subsection{Baselines}

We compare to the following methods:
\begin{itemize}
    \item \textbf{AdaFuse}~\cite{meng2021adafuse} trains a lightweight policy network to adaptively compute (or skip) feature map channels for each frame during inference. 
    \cc{The policy and recognition networks are trained jointly to optimize for both efficiency and performance.} We use the $\text{AdaFuse}^{\text{TSN}}_{\text{R50}}$ model with the provided hyper-parameters. 
    
    \item \textbf{STTS}~\cite{wang2021efficient} trains a module to rank spatio-temporal tokens derived from videos in a transformer-based model, and selects only the top-K tokens to speed up inference.
    
    \item \textbf{ListenToLook}~\cite{gao2020listentolook}: 
    uses the audio-based feature distillation module from \cite{gao2020listentolook} following the same audio processing and model architecture. 
\end{itemize}
\input{latex/tables/ablation_study}

These methods represent recent advances in efficient video recognition models. AdaFuse represents state-of-the-art approaches that achieve efficiency by reducing temporal redundancy in CNN models. STTS is one of the most recent approaches that efficiently reduces both spatial and temporal redundancy in ViT models, which achieves the state-of-the-art on Kinectics-400~\cite{carreira2017quo}. ListenToLook also relies on distillation, but using audio rather than head motion. 
For each model we generate multiple versions with different computation budgets
to plot accuracy vs.~GFLOPs.
We train all AdaFuse and STTS models with 4 input frames to align with the maximum frames used by our model. 
\KGnote{still need to clarify: fig 3 varies budget for STTS but not AdaFuse; how? and why no variation for AdaFuse?} 
\ST{For AdaFuse, we use the only provided hyper-parameter in the paper.\footnote{Modifying hyper-parameters to control the accuracy-efficiency trade-off results in unstable training and unreliable performance.} For STTS, we use three provided variants: $\text{T}_{0.5}^0$-$\text{S}_{0.7}^4$, $\text{T}_{0.8}^0$-$\text{S}_{0.9}^4$ and the full model without token selection.}
For ListenToLook we adopt the same efficiency-accuracy trade-off as our method, i.e., \KGnew{varying} the number of input frames.

In addition, we test variants of our method: 
\begin{itemize}
    \item \textbf{VisOnly-Distill} is our model without the IMU branch and fusion layer but trained with the same loss function. Performance of this model reveals the role
    of IMU in the process of distillation.

    \item \textbf{VisIMU} is our model trained with only $\mathcal{L}_\text{GT}$ in Equation~\ref{eq:gt}.
    It shows the effectiveness of  distillation from the video model compared with directly training the features with action labels. 
    
    \item \textbf{VisOnly} is an image-only model trained with $\mathcal{L}_\text{GT}$, which serves as the baseline.
\end{itemize}

\subsection{Main Results}

\input{latex/tables/model_architecture}
\textbf{Importance of IMU-guided distillation.}
Figure~\ref{fig:main_curve} shows the accuracy vs.~efficiency curves.
Methods towards the top-left of the plot represent those with both high accuracy and efficiency.
Our method achieves good accuracy  with low computational cost. 
Specifically, on \EPK, when $N=1$, \MODELNAME~improves over VisOnly-Distill by $8.4\%$ with only a small increase in computation. This result shows the effectiveness of IMU for reconstructing egocentric video features.
Compared to VisIMU, \MODELNAME~improves
by $9.9\%$, showing the effectiveness of knowledge distillation from the video model. Importantly, this reveals that \MODELNAME~does not simply benefit from the extra IMU context; our idea to approximate video features is necessary for best results. We see similar results on Ego4D.

\textbf{Comparison with the state of the art.}
Figure~\ref{fig:main_curve} also shows that \MODELNAME~achieves better accuracy with less computation than existing efficient video recognition models AdaFuse~\cite{meng2021adafuse},  STTS~\cite{wang2021efficient}, and ListenToLook~\cite{gao2020listentolook}.
With $N=4$ \TN{frames}, \MODELNAME~surpasses STTS by $7.4\%$ and AdaFuse by $4.2\%$ on \EPK, with $2 \times$ fewer GFLOPs, and surpasses both methods by $2.1\%$ on Ego4D.
In addition, \MODELNAME~surpasses ListenToLook by $7.4\%$ and $2.9\%$ on \EPK~and Ego4D respectively, which suggests that
head motion is more informative than audio for feature reconstruction in egocentric video.

\subsection{Analysis}

\textbf{Model component ablations.} Table~\ref{tab:ablation} ablates different design choices in our model, setting $N=1$ for all experiments.
We observe that training \MODELNAME~without $\mathcal{L}_1$, $\mathcal{L}_{\text{KD}}$ or $\mathcal{L}_{\text{GT}}$ deteriorates performance. 
Specifically, training without $\mathcal{L}_{\text{KD}}$ leads to the largest performance drop, which indicates that knowledge distillation is an essential component in our approach. 
Training without $\mathcal{L}_{1}$ also leads to a significant performance drop, which shows the importance of our idea to align 
features from the different modalities.
Further, our self-supervised pretraining stage 
is very effective at training the IMU extractor to encode useful motion information that is consistent with visual feature change. 
\cc{Both $\mathcal{L}_{\text{NCE}}$ and $\mathcal{L}_{1}$ contribute significantly to the final performance.}
Finally, we compare with a model that simply does multi-modal recognition with IMU (top row). The strong contrast here indicates the importance of our idea to use IMU to predict video model features, as opposed to simply adding IMU as an additional input modality.

\input{latex/figures/split}
\input{latex/tables/runtime}

\textbf{Impact of teacher video model architecture.}
In our main experiments we use MotionFormer~\cite{patrick2021keeping} as the teacher video model due to its strong performance on egocentric video tasks. To emphasize the generality of our idea, we show the performance of \MODELNAME~with other video teacher architectures in Table~\ref{tab:model_ab_acc}.
Similar to the MotionFormer model, we train these models on each of the labeled datasets, and then train our model using the resulting video models as the teacher.
As expected, better video teacher models lead to better student model performance.
More importantly, we observe consistent improvement by \MODELNAME~over the VisOnly-Distill baseline on both datasets and with different video teacher models, highlighting our idea's generality and versatility.

\input{latex/figures/quali_best_worse}
\input{latex/figures/retrival}
\input{latex/figures/quali_scene_motion}

\textbf{Where does our model work best/worst?}
In Figure~\ref{fig:main_curve} we saw that using IMU leads to an overall performance improvement on action recognition, indicating better video feature prediction capability. 
Next, we explore what kinds of clips are better reconstructed using \MODELNAME.
Figure~\ref{fig:split_analysis} shows the improvement of \MODELNAME~over the VisOnly-Distill model on 
Ego4D and
\EPK~split by action class.
We observe that IMU is more useful for actions with predictable head motion (e.g., \textit{break}, \textit{cut}, \textit{close}), and is less helpful for actions where head motion may be small or unrelated (e.g., \textit{empty}, \textit{fill}, \textit{press}).

Figure~\ref{fig:quali_bw} 
shows clip examples whose video features are best and worst reconstructed.
We observe that the best reconstructed clips (top) contain 
moderate head motion that is predictive of scene motion and action semantics. For example, the camera wearer's head moves slightly backwards while opening the cabinet.
On the other hand, more poorly reconstructed clips tend to contain little head motion (third row)---in which case IMU is redundant to the RGB frame---or drastic head motion that is weakly correlated with the camera wearer's activity and introduces blur to the frame (last row). 

\textbf{Efficiency analysis.}
To compare the efficiency of different models, aside from GFLOPs, we also compare their inference run-time and number of parameters. 
For run-time, we record the time spent to infer a single video clip's label with a single A40 GPU, and take the average time over the full validation datasets of Ego4D and \EPK~with batch-size of 32. Table~\ref{tab:runtime} shows the results.
\MODELNAME~runs much faster than the other methods. Notably, it reduces the GFLOPs of MotionFormer by nearly 200$\times$. 
Furthermore, it runs $6.5 \times$ faster than STTS~\cite{wang2021efficient} while achieving $4.4\%$ higher accuracy on \EPK. 

\subsection{Qualitative Results}
\textbf{What do \MODELNAME~features capture?}
To explore this, we pair a single input frame with different IMU clips as inputs to \MODELNAME, then retrieve the nearest video clip for each resulting anticipated video feature. 
Figure~\ref{fig:retrival} illustrates this.
We see that \MODELNAME~outputs video features that all involve interaction with the cabinet (right panel), and is able to use different IMU inputs to retrieve different video clips that show consistent camera motion. In contrast,
VisOnly-Distill only retains the semantic context to retrieve a single clip. These results indicate that \MODELNAME~is able to approximate video features that capture both semantic and motion information. 
See Supp.~for more (and animated) results.

\textbf{Is there evidence \MODELNAME~captures scene motion?} 
Figure~\ref{fig:quali_scene_motion} shows how our features learned with \emph{head} motion can nonetheless expose certain scene motion cues.
\MODELNAME~improves the accuracy over VisOnly-Distill on ambiguous categories  (like \textit{close} and \textit{put}) by a large margin ($20.3\%$ and $10.4\%$ on \EPK, $8.5\%$ and $3.9\%$ on Ego4D).
See caption for details.

%% file: latex/figures/main_curve.tex
\begin{figure*}[t]
    \centering
    
    \includegraphics[width=1.00\linewidth]{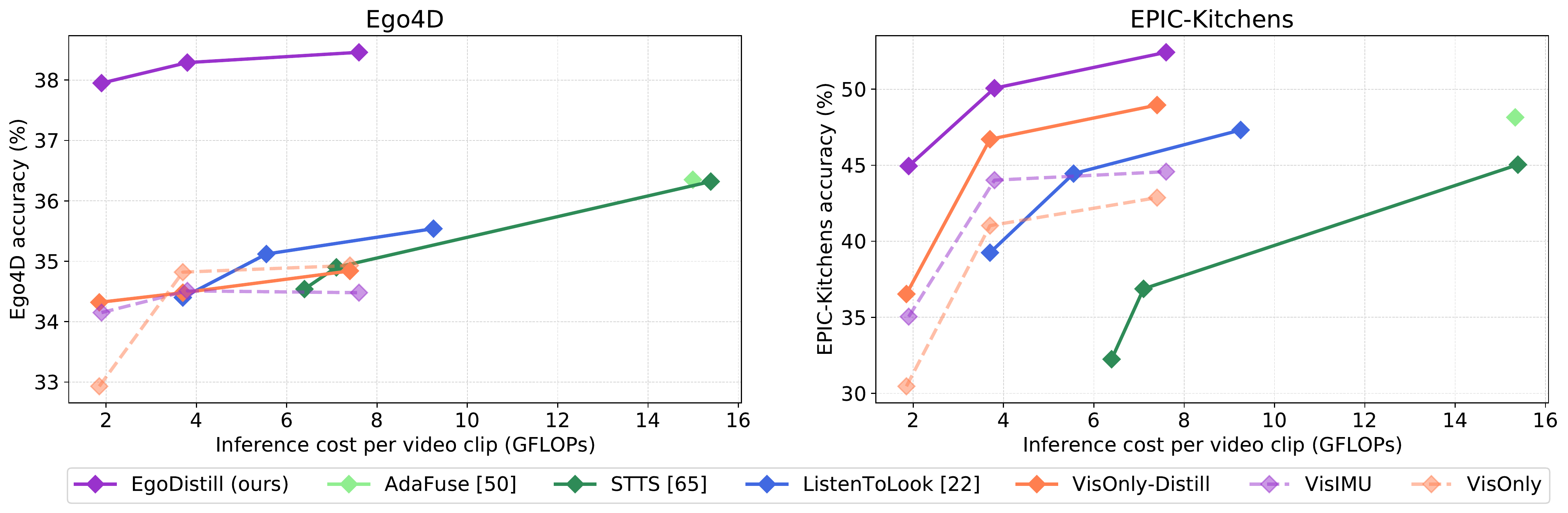}
  
    \vspace{-0.8em}
    \caption{\textbf{Accuracy vs. efficiency for action recognition on Ego4D (left) and \EPK~ (right).} \MODELNAME~outperforms state-of-the-art efficient video recognition methods that adaptively sample video content, while using 4$\times$ to 8$\times$ fewer GFLOPs.
    }  \label{fig:main_curve}
    
\vspace{-1.2em}
\end{figure*}

%% file: latex/tables/ablation_study.tex
\begin{table}[t]\centering
\scriptsize
\setlength\tabcolsep{5pt}
\begin{tabular}{ccccccc}\toprule
$\mathcal{L}_{\text{KD}}$ & $\mathcal{L}_1$ & $\mathcal{L}_{\text{GT}}$ & $\mathcal{L}_1$-pretrain & $\mathcal{L}_{\text{NCE}}$-pretrain & Ego4D & \EPK \\\cmidrule{1-7}
 &  & \checkmark &  &  & 34.15 & 35.04 \\
\midrule
 & \checkmark & \checkmark & \checkmark & \checkmark & 35.51 & 39.33 \\
\checkmark &  & \checkmark & \checkmark & \checkmark & 37.71 & 42.20 \\
\checkmark & \checkmark &  & \checkmark & \checkmark & 37.46 & 43.17 \\
\midrule
\checkmark & \checkmark & \checkmark & & & 36.99 & 41.21 \\
\checkmark & \checkmark & \checkmark &  & \checkmark & 37.26 & 42.30 \\
\checkmark & \checkmark & \checkmark & \checkmark &  & 37.49 & 43.51 \\
\midrule
\checkmark & \checkmark & \checkmark & \checkmark & \checkmark & \textbf{37.95} & \textbf{44.95} \\
\bottomrule
\end{tabular}
\caption{
\textbf{Ablation study of model components.} We compare the accuracy of \MODELNAME~with different components under $N=1$.
}
\label{tab:ablation}
\vspace{-1.8em}
\end{table}

%% file: latex/tables/model_architecture.tex
\begin{table}[t] 
\centering
\setlength\tabcolsep{3pt}
\scriptsize
\begin{tabular}{lccccccc}\toprule
\multirow{2}{*}{Source Model} &\multicolumn{3}{c}{Ego4D} &\multicolumn{3}{c}{\EPK} \\
 & Video &\MODELNAME & VisOnly-D & Video & \MODELNAME & VisOnly-D \\\cmidrule{1-8} 
MFormer~\cite{patrick2021keeping} & 46.38 & \textbf{37.95} & 34.32 & 77.28 & \textbf{44.95} & 37.20 \\
MViT~\cite{fan2021multiscale} & 40.32 & \textbf{36.46} & 33.40 & 53.38 & \textbf{36.90} & 31.22 &\\
SlowFast~\cite{9008780} & 40.52 & \textbf{33.29} & 33.04 & 58.34 & \textbf{39.42} & 33.47 \\
X3D~\cite{Feichtenhofer2020X3DEA} & 37.56 & \textbf{33.57} & 32.90 & 52.28 & \textbf{36.34} & 31.71 &\\

\bottomrule
\end{tabular}
\caption{\textbf{Versatility to model architectures.} \MODELNAME~outperforms the baseline for multiple common architectures, showing the generality of our idea.  ``Video" refers to the more expensive source model. We show the model accuracy under $N=1$. }\label{tab:model_ab_acc}
\end{table}

%% file: latex/figures/split.tex
\begin{figure}[t]
  \centering
  \includegraphics[width=1.0\linewidth]{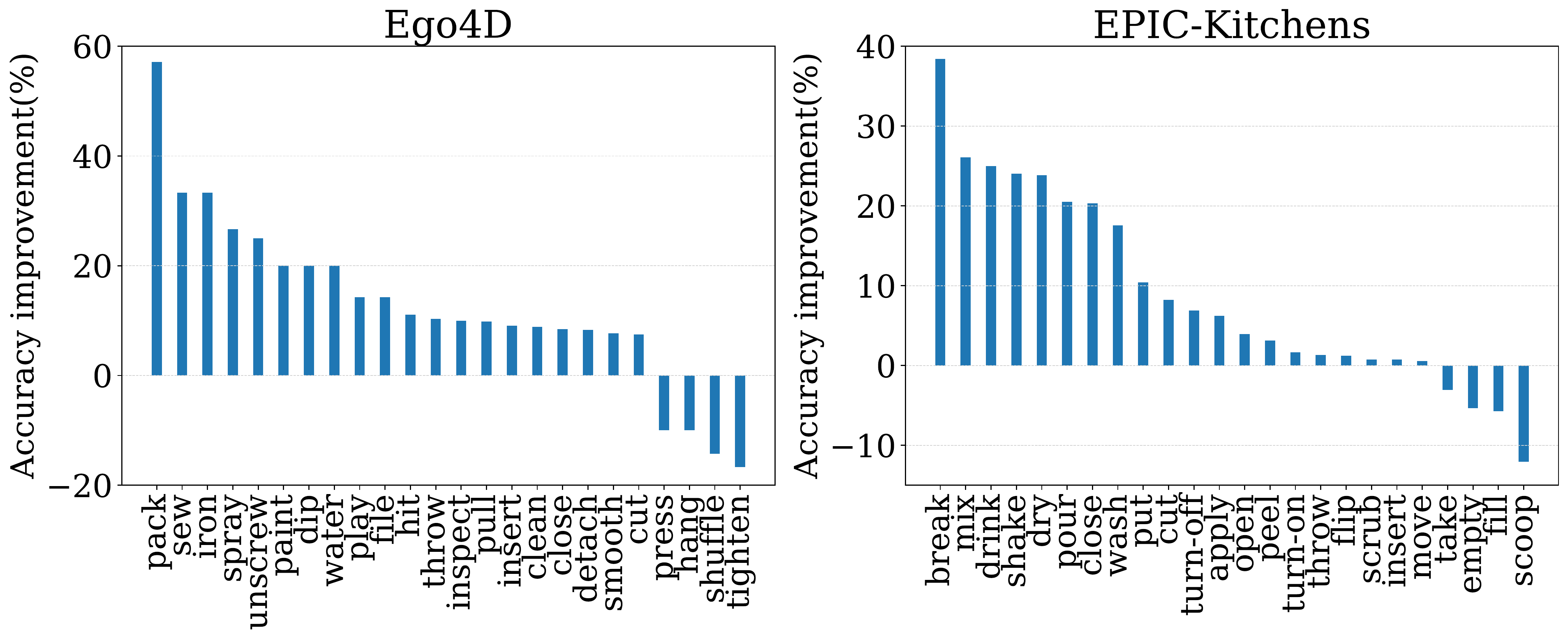}
  \caption{\textbf{Per-class accuracy improvement} over VisOnly-Distill. Best and worst performing classes are shown.
   }
   \label{fig:split_analysis}
\end{figure}

%% file: latex/tables/runtime.tex
\begin{table}[t]
\setlength\tabcolsep{8pt}
\scriptsize
\centering
\begin{tabular}{lccc}
\toprule
& GFLOPs  & Runtime (ms) & Parameters (M) \\
\midrule
Video~\cite{patrick2021keeping}         & 369.51 & 10.70 & 108.91 \\
AdaFuse~\cite{meng2021adafuse}          & 15.20 & 2.04 & 38.85 \\
STTS~\cite{wang2021efficient}           & 7.19 & 1.63 & 36.63 \\
ListenToLook~\cite{gao2020listentolook} & 3.10 & 0.43 & 25.53 \\
\MODELNAME                               & \textbf{1.91} & \textbf{0.25} & \textbf{20.56} \\
\bottomrule
\end{tabular}
\vspace{-0.1in}
\caption{\textbf{Efficiency analysis.} Our approach is the most efficient. ``Video" refers to the original (full-clip) feature. Lower is better.}
\label{tab:runtime}
\end{table}

%% file: latex/figures/quali_best_worse.tex
\begin{figure}[t!]
  \centering
  \includegraphics[width=1.0\linewidth]{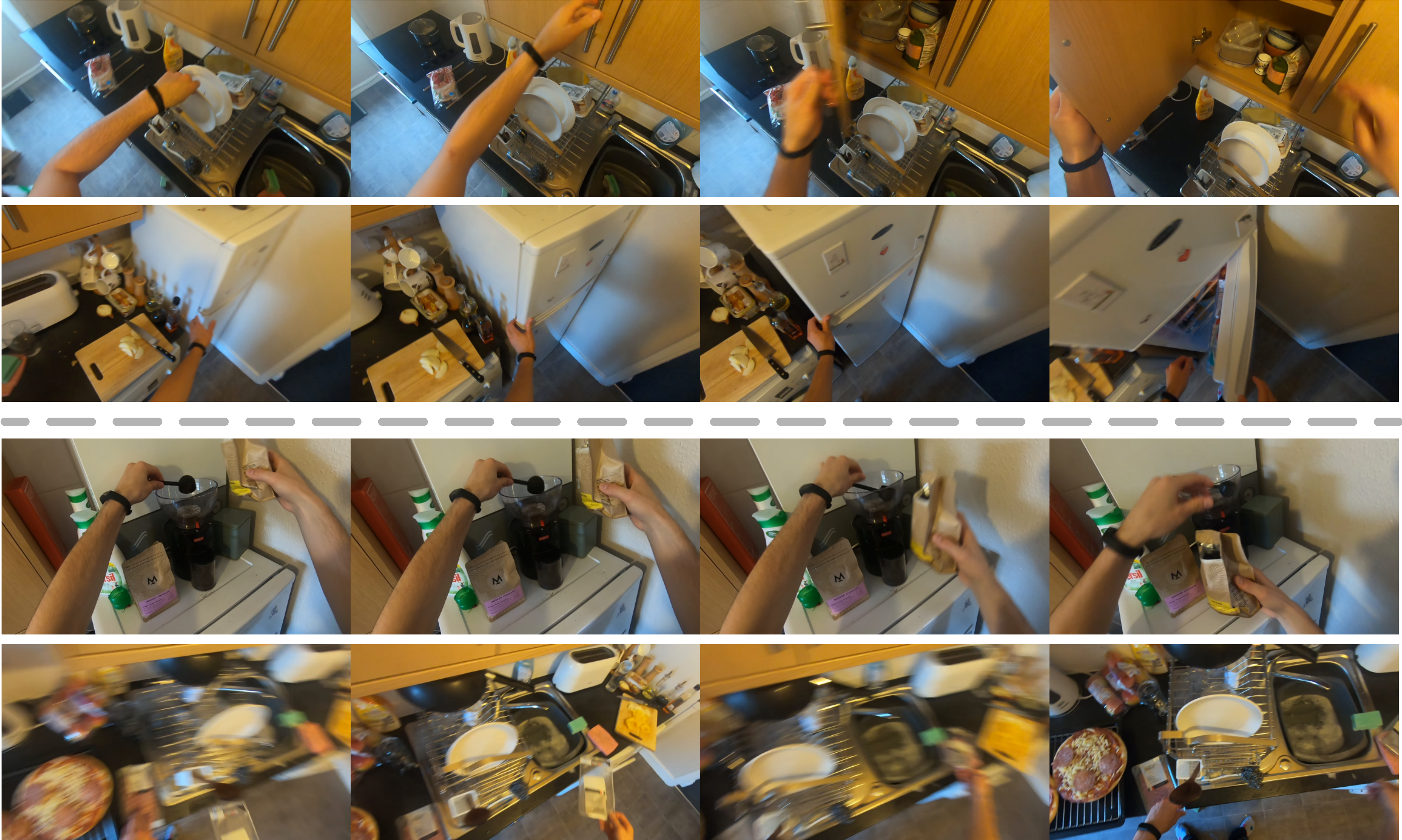}
  \caption{\textbf{Best (top) and worst (bottom) reconstructed videos.}} 
   \label{fig:quali_bw}
\end{figure}

%% file: latex/figures/retrival.tex
\begin{figure*}[t]
  \centering
\vspace*{-0.2in}
\includegraphics[width=1.0\linewidth]{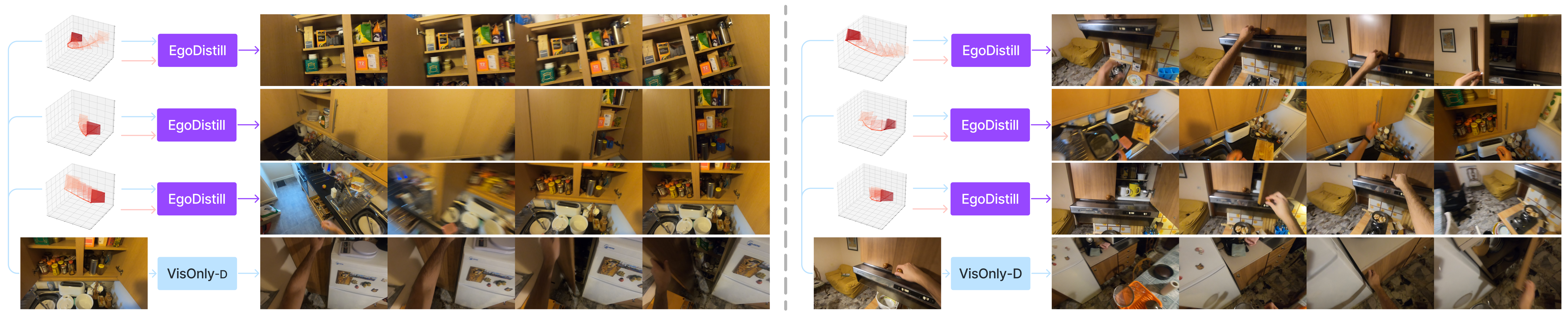}
\vspace*{-0.3in}
\caption{
\textbf{Retrieving video clips with \MODELNAME.}
Given a query frame (bottom left) and a paired IMU segment (red camera frustums) 
, we retrieve the nearest clip in the video dataset according to \MODELNAME~and visualize its (unobserved) frames (strip to the right). 
Compared to VisOnly-Distill, which outputs a single feature for a given input frame (bottom row), \MODELNAME~outputs a distinct 
feature by conditioning on IMU, showing its ability to preserve 
both semantic and motion during reconstruction. 
For instance, in the top-right example, \MODELNAME~retains the cabinet interaction semantics in the frame as well as the upward camera-motion in the IMU. 
Zoom in to view best.}

   \label{fig:retrival}
\end{figure*}

%% file: latex/figures/quali_scene_motion.tex
\begin{figure*}[t]
  \centering
  \includegraphics[width=1.0\linewidth]{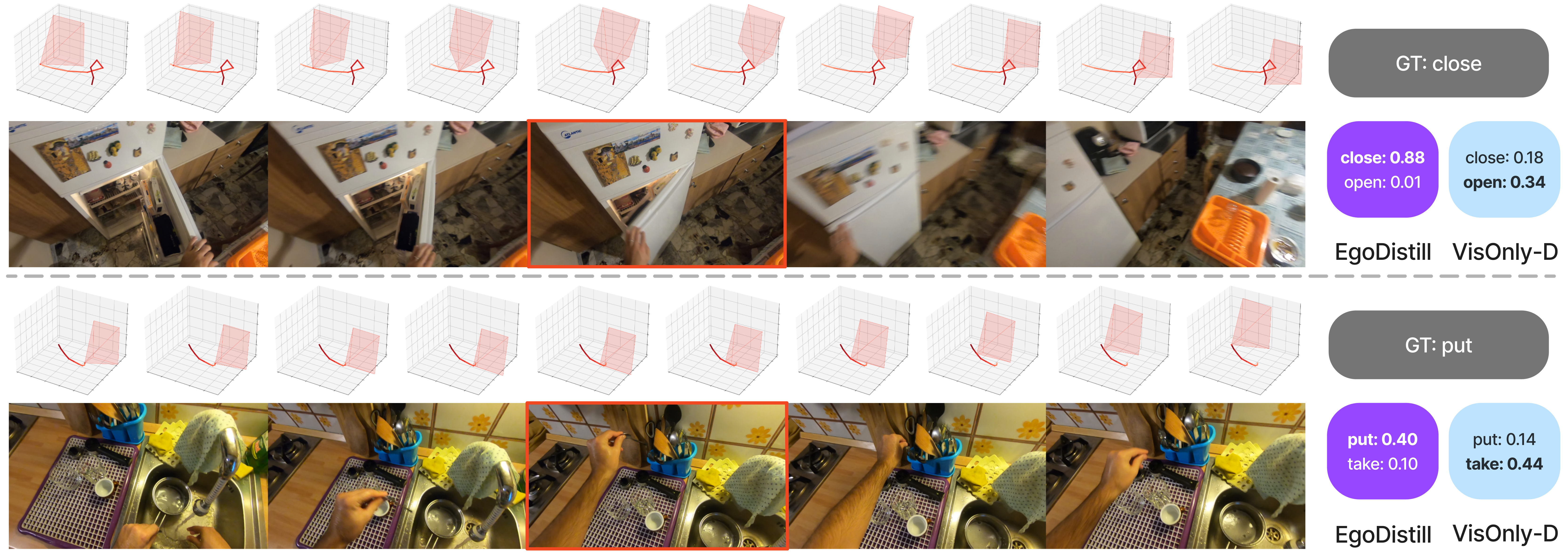}
  \vspace*{-0.3in}
  \caption{\textbf{Anticipating scene motion with \MODELNAME.} For each clip, we show the head motion and video frames. Note, only the center frame (red border) is observed by the model. 
  Action classification scores are shown on the right. 
  \MODELNAME~can successfully anticipate scene motion and disambiguate the action semantics in the input frame. For example, in the top center frame, the image alone cannot reveal if the door is being opened or closed, whereas our feature, learned with head motion, recovers correlations with the \emph{scene motion} (i.e., hand motion and door motion) to disambiguate ``close" from ``open". 
  A similar effect for ``put" vs.~``take" is seen in the second example.
   }
   \label{fig:quali_scene_motion}
\end{figure*}

%% file: latex/sections/conclusion.tex
We present \MODELNAME, the first model to explore egocentric video feature approximation for fast recognition.
Experiments on action recognition on Ego4D and \EPK~ demonstrate that our model achieves a good balance between accuracy and efficiency, outperforming state-of-the-art efficient video understanding methods. 
Our approach has great potential to accelerate video understanding for egocentric videos using a data stream that is already ubiquitous in egocentric cameras. 
\ST{In the future, we plan to investigate how to use head motion for long-term human activity understanding with room context 
and visual correspondence learning for multi-view videos.}

%% file: latex/supp.tex
\appendix

The supplementary materials of this work consist of:

\begin{enumerate} [label=\Alph*.]
    \item Supplementary video.
    \item Dataset details.
    \item Implementation details.
    \item Additional analysis of our model.
\end{enumerate}

\section{Supplementary Video}
In our supplementary video, we have a brief introduction of our work. More importantly, we show animated videos of Best and Worse reconstructed clips (Figure~\ref{fig:quali_bw}), Retrieving video clips with EgoDistill (Figure~\ref{fig:retrival}), and Anticipating scene motion with EgoDistill (Figure~\ref{fig:quali_scene_motion}).

Animated version of these figures better show head motion and video dynamics. 
We recommend viewing the supplementary video for better understanding of our method and results.

\section{Dataset Details.}
We use two datasets in our experiments: Ego4D~\cite{Ego4D2022CVPR} and \EPK-100~\cite{Damen2022RESCALING}.
In this section we describe more details about how we create our training and evaluation data.

\begin{enumerate}
    \item \textbf{Ego4D}~\cite{Ego4D2022CVPR} contains 3,670 hours of egocentric videos of people performing diverse tasks (from cooking to farming) across the globe. As action recognition is not part of the original 
    Ego4D benchmark, we construct this task with annotations from the Hands+Objects temporal localization benchmark~\cite{Ego4D2022CVPR}.
    Specifically, for each hand-objects interaction temporal annotation, we take the video clip between the pre-frame and post-frame of the annotation as input, and use the annotated verb for this interaction as label.

    We include clips with paired IMU and audio, and consider classes with at least 2 labeled instances, resulting in 94 action categories with 12.1k videos in total.
    In average, each clip has 2.2 second duration. Then, we randomly split data from each category into training and evaluation sets with 70\%:30\% ratio.
    Finally, we obtain a 94-class action recognition dataset with 8.5k training videos and 3.6k evaluation videos. 
    
    \item \textbf{\EPK}~\cite{Damen2022RESCALING} contains 100 hours of egocentric videos capturing daily activities in kitchen environments. We use annotations from the action recognition benchmark in our experiment.
    
    We select videos that have paired IMU and audio data, and split the resulting data by camera-wearer, ensuring non-overlapping splits following the original benchmark setting.
    Specifically, we take videos captured by camera-wearer id starting with P30, P35, P37 as evaluation videos and use all the remaining videos as training videos.
    This results in a 62-class action dataset with 29k training videos and 6.2k evaluation videos. 

\end{enumerate}

\section{Implementation Details.}
\textbf{IMU input processing.} For each input clip, IMU input is a $422\times6$ matrix (around 2 seconds with 198Hz frequency), representing the accelerometer and gyroscope readings \TN{along the} $xyz$ axes. We observe that the raw IMU input has significant drifting and bias issues. This induces inconsistent correspondence between camera motion and IMU reading across different clips and videos. Therefore, for IMU reading of each clip, on each dimension we separately subtract raw readings by the mean values on the corresponding dimension. This operation normalizes IMU readings in each dimension to have zero average value. In this way, our model can only focus on the temporal motion patterns in each clip.

\textbf{Audio input processing.} For ListenToLook~\cite{gao2020listentolook}, we process the audio input in the same way mentioned in the paper. Specifically, we subsample the audio at 16kHZ, and compute STFT using Hann window size of 400 and hop length of 160. Please refer to~\cite{gao2020listentolook} for more details.

\textbf{Model architecture.} For the image backbone, we use the ImageNet-pretrained ResNet-18 model. For the IMU backbone, we use a 5-layer 1D Dilated CNN, as found effective for IMU data processing~\cite{brossard2020denoising}. We use the same network setting (kernel dimension, dilation gap and channel dimension) as in prior work~\cite{brossard2020denoising}. The feature fusion model consists of a concatenation operation following two fully-connected layers with hidden dimension of 1024. Each layer except for the output layer is followed by a ReLU activation. The output dimension is the same as the teacher video model's feature dimension (768 in the case of MotionFormer). When $N>1$, we use a one-layer GRU module to aggregate extracted features for each frame. We use a single-directioal GRU with hidden dimension of 512.

\textbf{Model training.} We train our models in two stages. In the self-supervised IMU feature learning stage, we train random initialized IMU encoder $f_\mathcal{M}$, IMU predictor $h$ and the fusion network $\Pi$ with $\mathcal{L}_{\text{NCE}}$. Here the image encoder $f_\mathcal{I}$ is a fixed ImageNet pretrained model. On both datasets, we train the model for 50 epochs with AdamW and batch size 64. The initial training rate is $1e^{-4}$. We decay the training rate by $0.1$ at epoch $30$ and epoch $40$. In the second video feature distillation stage, we initialize the model with parameters obtained in the last stage and finetune. On both datasets, we use AdamW with batch size 64 and initial learning rate $1e^{-4}$. On Ego4D, we train for 150 epochs. We decay the training rate by $0.1$ at epoch $90$ and epoch $120$. On \EPK, we train for 50 epochs. We decay the training rate by $0.1$ at epoch $30$ and epoch $40$.

\input{latex/tables/frame_selection}

\section{Analysis.}
\textbf{Effect of frame selection.} In Section 3.2, we mentioned that we use uniform sampling to obtain the $N$ frames from each video clip. In this section, we compare the performance of our work under uniform sampling with other heuristics. Specifically, we compare with random sampling, the first $N$ frames, the last $N$ frames and the center $N$ frames. We show the results in Table~\ref{tab:frame} under $N=4$. These results indicate that uniform sampling leads to the best performance on average. Intuitively, uniform sampling on average leads to a broader coverage of both semantic contexts as well as scene motion.

\textbf{Why we set $N$ to be small.} In our experiments, we set $N$ to be 1 to 4. Using larger $N$ (\textit{e.g.,} 8 or 16) with densely sampled frames could lead to better results of all the methods with more computational cost. Efficient video understanding methods could benefit more as they have better temporal aggregation mechanisms given densely-sampled frames. However, the core purpose of our model is to deal with cases where we only use a few number of samples. Therefore, our model is not comparable to video clip models under dense-frame setting. 
Furthermore, setting $N$ to be a small number is very important in many applications. 
As loading more image frames takes additional time and memory, applications with streaming videos or low-resource AR/VR devices will benefit from loading only a few frames.




%% file: latex/tables/frame_selection.tex
\begin{table}[t]
\setlength\tabcolsep{15pt}
\centering
\begin{tabular}{lcc}
\toprule
& Ego4D  & \EPK\\
\midrule
uniform & 38.46 & 52.43\\
\midrule
random & 36.85 & 48.48 \\
first & 38.68 & 46.40 \\
last & 35.46 & 41.72 \\
center & 37.04 & 44.85 \\
\bottomrule
\end{tabular}
\vspace{-0.1in}
\caption{\textbf{Effect of frame selection.} We compare the accuracy of using different frame selection heuristics for EgoDistill when $N=4$. We observe that Uniform on average achieves better results.}
\label{tab:frame}
\end{table}

%% file: main.bbl
\begin{thebibliography}{10}\itemsep=-1pt

\bibitem{Arandjelovi2017LookLA}
Relja Arandjelovi{\'c} and Andrew Zisserman.
\newblock Look, listen and learn.
\newblock {\em 2017 IEEE International Conference on Computer Vision (ICCV)},
  pages 609--617, 2017.

\bibitem{10.5555/3157096.3157196}
Yusuf Aytar, Carl Vondrick, and Antonio Torralba.
\newblock Soundnet: Learning sound representations from unlabeled video.
\newblock In {\em Proceedings of the 30th International Conference on Neural
  Information Processing Systems}, NIPS'16, page 892–900, Red Hook, NY, USA,
  2016. Curran Associates Inc.

\bibitem{app10124213}
Anna Borowska-Terka and Pawel Strumillo.
\newblock Person independent recognition of head gestures from parametrised and
  raw signals recorded from inertial measurement unit.
\newblock {\em Applied Sciences}, 10(12), 2020.

\bibitem{Bortz1971ANM}
John~E. Bortz.
\newblock A new mathematical formulation for strapdown inertial navigation.
\newblock {\em IEEE Transactions on Aerospace and Electronic Systems},
  AES-7:61--66, 1971.

\bibitem{Brajdic2013WalkDA}
Agata Brajdic and Robert~K. Harle.
\newblock Walk detection and step counting on unconstrained smartphones.
\newblock {\em Proceedings of the 2013 ACM international joint conference on
  Pervasive and ubiquitous computing}, 2013.

\bibitem{brossard2020denoising}
M. {Brossard}, S. {Bonnabel}, and A. {Barrau}.
\newblock Denoising imu gyroscopes with deep learning for open-loop attitude
  estimation.
\newblock {\em IEEE Robotics and Automation Letters}, 5(3):4796--4803, 2020.

\bibitem{Cao_2022_CVPR}
Xiya Cao, Caifa Zhou, Dandan Zeng, and Yongliang Wang.
\newblock Rio: Rotation-equivariance supervised learning of robust inertial
  odometry.
\newblock In {\em Proceedings of the IEEE/CVF Conference on Computer Vision and
  Pattern Recognition (CVPR)}, pages 6614--6623, June 2022.

\bibitem{carreira2017quo}
Joao Carreira and Andrew Zisserman.
\newblock Quo vadis, action recognition? a new model and the kinetics dataset.
\newblock In {\em proceedings of the IEEE Conference on Computer Vision and
  Pattern Recognition}, pages 6299--6308, 2017.

\bibitem{7350781}
Chen Chen, Roozbeh Jafari, and Nasser Kehtarnavaz.
\newblock Utd-mhad: A multimodal dataset for human action recognition utilizing
  a depth camera and a wearable inertial sensor.
\newblock In {\em 2015 IEEE International Conference on Image Processing
  (ICIP)}, pages 168--172, 2015.

\bibitem{Cho2014LearningPR}
Kyunghyun Cho, Bart van Merrienboer, Çaglar G{\"u}lçehre, Dzmitry Bahdanau,
  Fethi Bougares, Holger Schwenk, and Yoshua Bengio.
\newblock Learning phrase representations using rnn encoder–decoder for
  statistical machine translation.
\newblock In {\em EMNLP}, 2014.

\bibitem{Damen2022RESCALING}
Dima Damen, Hazel Doughty, Giovanni~Maria Farinella, , Antonino Furnari, Jian
  Ma, Evangelos Kazakos, Davide Moltisanti, Jonathan Munro, Toby Perrett, Will
  Price, and Michael Wray.
\newblock Rescaling egocentric vision: Collection, pipeline and challenges for
  epic-kitchens-100.
\newblock {\em International Journal of Computer Vision (IJCV)}, 130:33–55,
  2022.

\bibitem{Datta_2022_CVPR}
Samyak Datta, Sameer Dharur, Vincent Cartillier, Ruta Desai, Mukul Khanna,
  Dhruv Batra, and Devi Parikh.
\newblock Episodic memory question answering.
\newblock In {\em Proceedings of the IEEE/CVF Conference on Computer Vision and
  Pattern Recognition (CVPR)}, pages 19119--19128, June 2022.

\bibitem{8730690}
Alexander Diete, Timo Sztyler, and Heiner Stuckenschmidt.
\newblock Vision and acceleration modalities: Partners for recognizing complex
  activities.
\newblock In {\em 2019 IEEE International Conference on Pervasive Computing and
  Communications Workshops (PerCom Workshops)}, pages 101--106, 2019.

\bibitem{ehsani2018dog}
Kiana Ehsani, Hessam Bagherinezhad, Joseph Redmon, Roozbeh Mottaghi, and Ali
  Farhadi.
\newblock Who let the dogs out? modeling dog behavior from visual data.
\newblock In {\em CVPR}, 2018.

\bibitem{ehsani2020learning}
Kiana Ehsani, Daniel Gordon, Thomas Nguyen, Roozbeh Mottaghi, and Ali Farhadi.
\newblock Learning visual representation from human interactions.
\newblock {\em International Conference on Learning Representations}, 2021.

\bibitem{fan2021multiscale}
Haoqi Fan, Bo Xiong, Karttikeya Mangalam, Yanghao Li, Zhicheng Yan, Jitendra
  Malik, and Christoph Feichtenhofer.
\newblock Multiscale vision transformers.
\newblock In {\em ICCV}, 2021.

\bibitem{Feichtenhofer2020X3DEA}
Christoph Feichtenhofer.
\newblock X3d: Expanding architectures for efficient video recognition.
\newblock {\em 2020 IEEE/CVF Conference on Computer Vision and Pattern
  Recognition (CVPR)}, pages 200--210, 2020.

\bibitem{9008780}
Christoph Feichtenhofer, Haoqi Fan, Jitendra Malik, and Kaiming He.
\newblock Slowfast networks for video recognition.
\newblock In {\em 2019 IEEE/CVF International Conference on Computer Vision
  (ICCV)}, pages 6201--6210, 2019.

\bibitem{Feichtenhofer2016ConvolutionalTN}
Christoph Feichtenhofer, Axel Pinz, and Andrew Zisserman.
\newblock Convolutional two-stream network fusion for video action recognition.
\newblock {\em 2016 IEEE Conference on Computer Vision and Pattern Recognition
  (CVPR)}, pages 1933--1941, 2016.

\bibitem{7557075}
Christian Forster, Luca Carlone, Frank Dellaert, and Davide Scaramuzza.
\newblock On-manifold preintegration for real-time visual--inertial odometry.
\newblock {\em IEEE Transactions on Robotics}, 33(1):1--21, 2017.

\bibitem{9009049}
C. Gan, H. Zhao, P. Chen, D. Cox, and A. Torralba.
\newblock Self-supervised moving vehicle tracking with stereo sound.
\newblock In {\em 2019 IEEE/CVF International Conference on Computer Vision
  (ICCV)}, pages 7052--7061, Los Alamitos, CA, USA, nov 2019. IEEE Computer
  Society.

\bibitem{gao2020listentolook}
Ruohan Gao, Tae-Hyun Oh, Kristen Grauman, and Lorenzo Torresani.
\newblock Listen to look: Action recognition by previewing audio.
\newblock In {\em Proceedings of the IEEE/CVF Conference on Computer Vision and
  Pattern Recognition}, pages 10457--10467, 2020.

\bibitem{Garcia_2018_ECCV}
Nuno~C. Garcia, Pietro Morerio, and Vittorio Murino.
\newblock Modality distillation with multiple stream networks for action
  recognition.
\newblock In {\em The European Conference on Computer Vision (ECCV)}, September
  2018.

\bibitem{ghodrati2021}
Amir Ghodrati, Babak~Ehteshami Bejnordi, and Amirhossein Habibian.
\newblock Frameexit: Conditional early exiting for efficient video recognition.
\newblock In {\em Proceedings of the IEEE Conference on Computer Vision and
  Pattern Recognition}, 2021.

\bibitem{girdhar2022omnivore}
Rohit Girdhar, Mannat Singh, Nikhila Ravi, Laurens van~der Maaten, Armand
  Joulin, and Ishan Misra.
\newblock {Omnivore: A Single Model for Many Visual Modalities}.
\newblock In {\em CVPR}, 2022.

\bibitem{Ego4D2022CVPR}
Kristen Grauman, Andrew Westbury, Eugene Byrne, Zachary Chavis, Antonino
  Furnari, Rohit Girdhar, Jackson Hamburger, Hao Jiang, Miao Liu, Xingyu Liu,
  Miguel Martin, Tushar Nagarajan, Ilija Radosavovic, Santhosh~Kumar
  Ramakrishnan, Fiona Ryan, Jayant Sharma, Michael Wray, Mengmeng Xu,
  Eric~Zhongcong Xu, Chen Zhao, Siddhant Bansal, Dhruv Batra, Vincent
  Cartillier, Sean Crane, Tien Do, Morrie Doulaty, Akshay Erapalli, Christoph
  Feichtenhofer, Adriano Fragomeni, Qichen Fu, Christian Fuegen, Abrham
  Gebreselasie, Cristina Gonzalez, James Hillis, Xuhua Huang, Yifei Huang,
  Wenqi Jia, Weslie Khoo, Jachym Kolar, Satwik Kottur, Anurag Kumar, Federico
  Landini, Chao Li, Yanghao Li, Zhenqiang Li, Karttikeya Mangalam, Raghava
  Modhugu, Jonathan Munro, Tullie Murrell, Takumi Nishiyasu, Will Price,
  Paola~Ruiz Puentes, Merey Ramazanova, Leda Sari, Kiran Somasundaram, Audrey
  Southerland, Yusuke Sugano, Ruijie Tao, Minh Vo, Yuchen Wang, Xindi Wu,
  Takuma Yagi, Yunyi Zhu, Pablo Arbelaez, David Crandall, Dima Damen,
  Giovanni~Maria Farinella, Bernard Ghanem, Vamsi~Krishna Ithapu, C.~V.
  Jawahar, Hanbyul Joo, Kris Kitani, Haizhou Li, Richard Newcombe, Aude Oliva,
  Hyun~Soo Park, James~M. Rehg, Yoichi Sato, Jianbo Shi, Mike~Zheng Shou,
  Antonio Torralba, Lorenzo Torresani, Mingfei Yan, and Jitendra Malik.
\newblock Ego4d: Around the {W}orld in 3,000 {H}ours of {E}gocentric {V}ideo.
\newblock In {\em IEEE/CVF Computer Vision and Pattern Recognition (CVPR)},
  2022.

\bibitem{GUITCHOUNTS2020512}
Grigori Guitchounts, Javier Masís, Steffen~B.E. Wolff, and David Cox.
\newblock Encoding of 3d head orienting movements in the primary visual cortex.
\newblock {\em Neuron}, 108(3):512--525.e4, 2020.

\bibitem{7780678}
Saurabh Gupta, Judy Hoffman, and Jitendra Malik.
\newblock Cross modal distillation for supervision transfer.
\newblock In {\em 2016 IEEE Conference on Computer Vision and Pattern
  Recognition (CVPR)}, pages 2827--2836, 2016.

\bibitem{pmlr-v9-gutmann10a}
Michael Gutmann and Aapo Hyvärinen.
\newblock Noise-contrastive estimation: A new estimation principle for
  unnormalized statistical models.
\newblock In Yee~Whye Teh and Mike Titterington, editors, {\em Proceedings of
  the Thirteenth International Conference on Artificial Intelligence and
  Statistics}, volume~9 of {\em Proceedings of Machine Learning Research},
  pages 297--304, Chia Laguna Resort, Sardinia, Italy, 13--15 May 2010. PMLR.

\bibitem{9196860}
Sachini Herath, Hang Yan, and Yasutaka Furukawa.
\newblock Ronin: Robust neural inertial navigation in the wild: Benchmark,
  evaluations, \& new methods.
\newblock In {\em 2020 IEEE International Conference on Robotics and Automation
  (ICRA)}, pages 3146--3152, 2020.

\bibitem{Hinton2015DistillingTK}
Geoffrey~E. Hinton, Oriol Vinyals, and Jeffrey Dean.
\newblock Distilling the knowledge in a neural network.
\newblock {\em NIPS Deep Learning Workshop}, abs/1503.02531, 2014.

\bibitem{huang2021what}
Yu Huang, Chenzhuang Du, Zihui Xue, Xuanyao Chen, Hang Zhao, and Longbo Huang.
\newblock What makes multi-modal learning better than single (provably).
\newblock In A. Beygelzimer, Y. Dauphin, P. Liang, and J.~Wortman Vaughan,
  editors, {\em Advances in Neural Information Processing Systems}, 2021.

\bibitem{IGNATOV2018915}
Andrey Ignatov.
\newblock Real-time human activity recognition from accelerometer data using
  convolutional neural networks.
\newblock {\em Applied Soft Computing}, 62:915--922, 2018.

\bibitem{jayaraman-iccv2015}
D. Jayaraman and K. Grauman.
\newblock {Learning image representations tied to egomotion}.
\newblock In {\em ICCV}, 2015.

\bibitem{5286542}
A.R. Jimenez, F. Seco, C. Prieto, and J. Guevara.
\newblock A comparison of pedestrian dead-reckoning algorithms using a low-cost
  mems imu.
\newblock In {\em 2009 IEEE International Symposium on Intelligent Signal
  Processing}, pages 37--42, 2009.

\bibitem{kazakos2019TBN}
Evangelos Kazakos, Arsha Nagrani, Andrew Zisserman, and Dima Damen.
\newblock Epic-fusion: Audio-visual temporal binding for egocentric action
  recognition.
\newblock In {\em IEEE/CVF International Conference on Computer Vision (ICCV)},
  2019.

\bibitem{kondratyuk2021movinets}
Dan Kondratyuk, Liangzhe Yuan, Yandong Li, Li Zhang, Mingxing Tan, Matthew
  Brown, and Boqing Gong.
\newblock Movinets: Mobile video networks for efficient video recognition.
\newblock In {\em Proceedings of the IEEE/CVF Conference on Computer Vision and
  Pattern Recognition}, pages 16020--16030, 2021.

\bibitem{10.5555/3327757.3327874}
Bruno Korbar, Du Tran, and Lorenzo Torresani.
\newblock Cooperative learning of audio and video models from self-supervised
  synchronization.
\newblock In {\em Proceedings of the 32nd International Conference on Neural
  Information Processing Systems}, NIPS'18, page 7774–7785, Red Hook, NY,
  USA, 2018. Curran Associates Inc.

\bibitem{Korbar2019SCSamplerSS}
Bruno Korbar, Du Tran, and Lorenzo Torresani.
\newblock Scsampler: Sampling salient clips from video for efficient action
  recognition.
\newblock {\em 2019 IEEE/CVF International Conference on Computer Vision
  (ICCV)}, pages 6231--6241, 2019.

\bibitem{inproceedings}
Masakatsu Kourogi and Takeshi Kurata.
\newblock A method of pedestrian dead reckoning for smartphones using frequency
  domain analysis on patterns of acceleration and angular velocity.
\newblock pages 164--168, 05 2014.

\bibitem{10.1145/1964897.1964918}
Jennifer~R. Kwapisz, Gary~M. Weiss, and Samuel~A. Moore.
\newblock Activity recognition using cell phone accelerometers.
\newblock {\em SIGKDD Explor. Newsl.}, 12(2):74–82, mar 2011.

\bibitem{10.1371/journal.pone.0075196}
Heike Leutheuser, Dominik Schuldhaus, and Bjoern~M. Eskofier.
\newblock Hierarchical, multi-sensor based classification of daily life
  activities: Comparison with state-of-the-art algorithms using a benchmark
  dataset.
\newblock {\em PLOS ONE}, 8(10):1--11, 10 2013.

\bibitem{Li_2022_CVPR}
Yiming Li, Ziang Cao, Andrew Liang, Benjamin Liang, Luoyao Chen, Hang Zhao, and
  Chen Feng.
\newblock Egocentric prediction of action target in 3d.
\newblock In {\em Proceedings of the IEEE/CVF Conference on Computer Vision and
  Pattern Recognition (CVPR)}, June 2022.

\bibitem{Liu_2022_CVPR}
Shaowei Liu, Subarna Tripathi, Somdeb Majumdar, and Xiaolong Wang.
\newblock Joint hand motion and interaction hotspots prediction from egocentric
  videos.
\newblock In {\em Proceedings of the IEEE/CVF Conference on Computer Vision and
  Pattern Recognition (CVPR)}, pages 3282--3292, June 2022.

\bibitem{Liu20213Dto2DDF}
Zhengzhe Liu, Xiaojuan Qi, and Chi-Wing Fu.
\newblock 3d-to-2d distillation for indoor scene parsing.
\newblock {\em 2021 IEEE/CVF Conference on Computer Vision and Pattern
  Recognition (CVPR)}, pages 4462--4472, 2021.

\bibitem{loshchilov2018fixing}
Ilya Loshchilov and Frank Hutter.
\newblock Fixing weight decay regularization in adam, 2018.

\bibitem{9680027}
Niall Lyons, Avik Santra, and Ashutosh Pandey.
\newblock Improved deep representation learning for human activity recognition
  using imu sensors.
\newblock In {\em 2021 20th IEEE International Conference on Machine Learning
  and Applications (ICMLA)}, pages 326--332, 2021.

\bibitem{9187232}
Sharmin Majumder and Nasser Kehtarnavaz.
\newblock Vision and inertial sensing fusion for human action recognition: A
  review.
\newblock {\em IEEE Sensors Journal}, 21(3):2454--2467, 2021.

\bibitem{meng2020ar}
Yue Meng, Chung-Ching Lin, Rameswar Panda, Prasanna Sattigeri, Leonid
  Karlinsky, Aude Oliva, Kate Saenko, and Rogerio Feris.
\newblock Ar-net: Adaptive frame resolution for efficient action recognition.
\newblock {\em arXiv preprint arXiv:2007.15796}, 2020.

\bibitem{meng2021adafuse}
Yue Meng, Rameswar Panda, Chung-Ching Lin, Prasanna Sattigeri, Leonid
  Karlinsky, Kate Saenko, Aude Oliva, and Rogerio Feris.
\newblock Adafuse: Adaptive temporal fusion network for efficient action
  recognition.
\newblock In {\em International Conference on Learning Representations}, 2021.

\bibitem{panda2021adamml}
Rameswar Panda, Chun-Fu Chen, Quanfu Fan, Ximeng Sun, Kate Saenko, Aude Oliva,
  and Rogerio Feris.
\newblock {AdaMML: Adaptive Multi-Modal Learning for Efficient Video
  Recognition}.
\newblock In {\em International Conference on Computer Vision (ICCV)}, 2021.

\bibitem{PARKER2020581}
Philip~R.L. Parker, Morgan~A. Brown, Matthew~C. Smear, and Cristopher~M. Niell.
\newblock Movement-related signals in sensory areas: Roles in natural behavior.
\newblock {\em Trends in Neurosciences}, 43(8):581--595, 2020.

\bibitem{Parker2022.02.01.478733}
Philip R.~L. Parker, Elliott T.~T. Abe, Emmalyn S.~P. Leonard, Dylan~M.
  Martins, and Cristopher~M. Niell.
\newblock Joint coding of visual input and eye/head position in v1 of freely
  moving mice.
\newblock {\em bioRxiv}, 2022.

\bibitem{patrick2021keeping}
Mandela Patrick, Dylan Campbell, Yuki~M. Asano, Ishan Misra~Florian Metze,
  Christoph Feichtenhofer, Andrea Vedaldi, and João~F. Henriques.
\newblock Keeping your eye on the ball: Trajectory attention in video
  transformers.
\newblock In {\em Advances in Neural Information Processing Systems (NeurIPS)},
  2021.

\bibitem{Plizzari_2022_CVPR}
Chiara Plizzari, Mirco Planamente, Gabriele Goletto, Marco Cannici, Emanuele
  Gusso, Matteo Matteucci, and Barbara Caputo.
\newblock E2(go)motion: Motion augmented event stream for egocentric action
  recognition.
\newblock In {\em Proceedings of the IEEE/CVF Conference on Computer Vision and
  Pattern Recognition (CVPR)}, pages 19935--19947, June 2022.

\bibitem{sigurdsson2018charades}
Gunnar~A Sigurdsson, Abhinav Gupta, Cordelia Schmid, Ali Farhadi, and Karteek
  Alahari.
\newblock Charades-ego: A large-scale dataset of paired third and first person
  videos.
\newblock {\em arXiv preprint arXiv:1804.09626}, 2018.

\bibitem{NIPS2014_00ec53c4}
Karen Simonyan and Andrew Zisserman.
\newblock Two-stream convolutional networks for action recognition in videos.
\newblock In Z. Ghahramani, M. Welling, C. Cortes, N. Lawrence, and K.Q.
  Weinberger, editors, {\em Advances in Neural Information Processing Systems},
  volume~27. Curran Associates, Inc., 2014.

\bibitem{Sun_2018_CVPR}
Shuyang Sun, Zhanghui Kuang, Lu Sheng, Wanli Ouyang, and Wei Zhang.
\newblock Optical flow guided feature: A fast and robust motion representation
  for video action recognition.
\newblock In {\em The IEEE Conference on Computer Vision and Pattern
  Recognition (CVPR)}, June 2018.

\bibitem{9744439}
Yin Tang, Lei Zhang, Fuhong Min, and Jun He.
\newblock Multi-scale deep feature learning for human activity recognition
  using wearable sensors.
\newblock {\em IEEE Transactions on Industrial Electronics}, pages 1--1, 2022.

\bibitem{Tao2021AttentionBasedSF}
Wenjin Tao, Haodong Chen, Md. Moniruzzaman, Ming~C. Leu, Zhaozheng Yi, and
  Ruwen Qin.
\newblock Attention-based sensor fusion for human activity recognition using
  imu signals.
\newblock {\em ArXiv}, abs/2112.11224, 2021.

\bibitem{10.1145/3494995}
Catherine Tong, Jinchen Ge, and Nicholas~D Lane.
\newblock Zero-shot learning for imu-based activity recognition using video
  embeddings.
\newblock {\em Proceedings of the ACM on Interactive, Mobile, Wearable and
  Ubiquitous Technologies}, 5(4):1--23, 2021.

\bibitem{Tran2018ACL}
Du Tran, Heng Wang, Lorenzo Torresani, Jamie Ray, Yann LeCun, and Manohar
  Paluri.
\newblock A closer look at spatiotemporal convolutions for action recognition.
\newblock {\em 2018 IEEE/CVF Conference on Computer Vision and Pattern
  Recognition}, pages 6450--6459, 2018.

\bibitem{Tsutsui2021HowYM}
Satoshi Tsutsui, Ruta Desai, and Karl Ridgeway.
\newblock How you move your head tells what you do: Self-supervised video
  representation learning with egocentric cameras and imu sensors.
\newblock {\em ICCV Workshop on Egocentric Perception, Interaction and
  Computing (EPIC)}, 2021.

\bibitem{Wang2019DeepLF}
Jindong Wang, Yiqiang Chen, Shuji Hao, Xiaohui Peng, and Lisha Hu.
\newblock Deep learning for sensor-based activity recognition: A survey.
\newblock {\em Pattern Recognit. Lett.}, 119:3--11, 2019.

\bibitem{wang2021efficient}
Junke Wang, Xitong Yang, Hengduo Li, Liu Li, Zuxuan Wu, and Yu-Gang Jiang.
\newblock Efficient video transformers with spatial-temporal token selection.
\newblock In {\em ECCV}, 2022.

\bibitem{wang2021adafocus}
Yulin Wang, Zhaoxi Chen, Haojun Jiang, Shiji Song, Yizeng Han, and Gao Huang.
\newblock Adaptive focus for efficient video recognition.
\newblock In {\em Proceedings of the IEEE/CVF International Conference on
  Computer Vision (ICCV)}, October 2021.

\bibitem{wang2022adafocusv2}
Yulin Wang, Yang Yue, Yuanze Lin, Haojun Jiang, Zihang Lai, Victor Kulikov,
  Nikita Orlov, Humphrey Shi, and Gao Huang.
\newblock Adafocus v2: End-to-end training of spatial dynamic networks for
  video recognition.
\newblock In {\em Proceedings of the IEEE/CVF Conference on Computer Vision and
  Pattern Recognition (CVPR)}, 2022.

\bibitem{s20102905}
Haoran Wei, Pranav Chopada, and Nasser Kehtarnavaz.
\newblock C-mhad: Continuous multimodal human action dataset of simultaneous
  video and inertial sensing.
\newblock {\em Sensors}, 20(10), 2020.

\bibitem{8994060}
Haoran Wei and Nasser Kehtarnavaz.
\newblock Simultaneous utilization of inertial and video sensing for action
  detection and recognition in continuous action streams.
\newblock {\em IEEE Sensors Journal}, 20(11):6055--6063, 2020.

\bibitem{Yan2018RIDIRI}
Hang Yan, Qi Shan, and Yasutaka Furukawa.
\newblock Ridi: Robust imu double integration.
\newblock In {\em ECCV}, 2018.

\bibitem{Yang2022EfficientDV}
Mingyu Yang, Yu Chen, and Hun-Seok Kim.
\newblock Efficient deep visual and inertial odometry with adaptive visual
  modality selection.
\newblock {\em ArXiv}, abs/2205.06187, 2022.

\bibitem{ECO_eccv18}
Mohammadreza Zolfaghari, Kamaljeet Singh, and Thomas Brox.
\newblock {ECO:} efficient convolutional network for online video
  understanding.
\newblock In {\em ECCV}, 2018.

\end{thebibliography}
